\documentclass[letterpaper, 10 pt, conference]{ieeeconf}  

\IEEEoverridecommandlockouts                              

\usepackage{graphics} 
\usepackage{epsfig} 
\usepackage{amsmath} 
\usepackage{amssymb}  
\usepackage{algorithm}
\usepackage{algpseudocode}
\usepackage{mathrsfs}
\usepackage{cite}
\usepackage{hyperref}
\usepackage{threeparttable}
\usepackage[T1]{fontenc}

\title{\LARGE \bf
MAN: Multi-Action Networks Learning
}

\author{Keqin Wang$^{1}$ Alison Bartsch$^{1}$ Amir Barati Farimani$^{1}$
\thanks{$^{1}$Department of Mechanical Engineering, Carnegie Mellon University, United States}
}

\begin{document}

\maketitle
\thispagestyle{empty}
\pagestyle{empty}

\begin{abstract}
Learning control policies with large discrete action spaces is a challenging problem in the field of reinforcement learning due to present inefficiencies in exploration. With high dimensional action spaces, there are a large number of potential actions in each individual dimension over which policies would be learned. In this work, we introduce a Deep Reinforcement Learning (DRL) algorithm call Multi-Action Networks (MAN) Learning that addresses the challenge of high-dimensional large discrete action spaces. We propose factorizing the N-dimension action space into N 1-dimensional components, known as sub-actions, creating a Value Neural Network for each sub-action. Then, MAN uses temporal-difference learning to train the networks synchronously, which is simpler than training a single network with a large action output directly. To evaluate the proposed method, we test MAN on three scenarios: an n-dimension maze task, a block stacking task, and then extend MAN to handle 12 games from the Atari Arcade Learning environment with 18 action spaces. Our results indicate that MAN learns faster than both Deep Q-Learning and Double Deep Q-Learning, implying our method is a better performing synchronous temporal difference algorithm than those currently available for large discrete action spaces.
\end{abstract}

\section{Introduction}

Learning a policy with a large discrete action space is a current critical challenge for high-dimensional control tasks. Specifically in reinforcement learning (RL), the exponential large action spaces impede performance. In particular, DQN \cite{mnih2013playing}\cite{mnih2015human}, and other similar algorithms have difficulty handling large action spaces. Unfortunately, large discrete action spaces are very prevalent, e.g. a control task in continuous action space can by solved by discretizing the actions creating a large discrete action space. Moreover, some learning pipelines may use images, where each image pixel is an action. These problems are challenging due to the large exploration space, i.e. there are numerous actions to choose from.

\begin{figure}[thpb]
  \centering
  \includegraphics[width=0.45\textwidth]{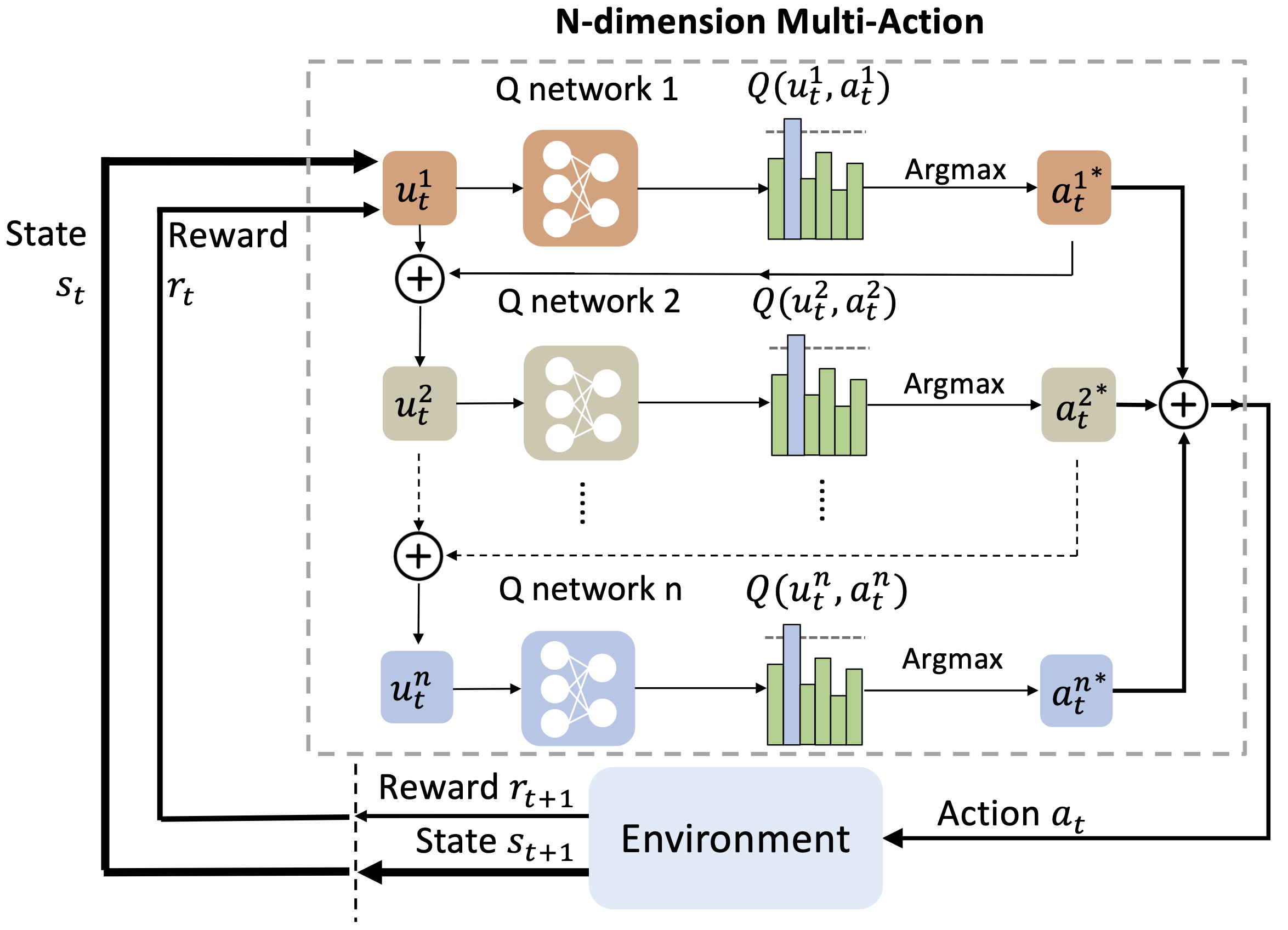}
  \caption{n-dimension MAN structure schematic. The MAN algorithm firstly splits the original action $a_t$ into $n$ sub-actions $(a^{1}_t,a^{2}_t,a^{3}_t, ... a^{n}_t)$; and then uses $n$ streams to separately estimate values of sub-actions; finally combines these sub-actions back to the original action. Both networks output Q-values for each sub-action. Note, viewing the content in the dashed box as a whole is the traditional reinforcement learning pipeline.}
  \label{pipeline}
\end{figure}

Currently, one of the common methods developed to address the challenge of large action spaces is to reduce the dimension of the action space by introducing features and approximating actions \cite{dulac2015deep}. These methods requires additional models to map between features and actions, which is both time consuming and introduces new error into the mappings. Another famous approach is to learn the policy directly, e.g. policy-gradient methods \cite{sutton1999policy}. However, the computations in policy-gradient methods are more complicated than value-iteration methods. For now, some recent works define certain decompositions in an effort to deal with high-dimensional action spaces. Among them, Factored Actions Reinforcement Learning (FARL) in \cite{pierrotfactored}, which uses several neural networks to predict one dimension at a time to simplify complex functions over high-dimensional spaces, significantly motivated our paper. This framework has reached some astonishing successes such as in Video Games StarCraft and DOTA 2 \cite{berner2019dota}.

In this paper, we present a novel DRL algorithm known as \textit{Multi-Action Networks} (MAN) Learning, leveraging DQN, the powerful capabilities of temporal difference (TD) learning \cite{sutton1988learning} and the FARL framework. Our method addresses the challenge of large discrete action spaces by splitting the original n-dimension large action space into n smaller sub-action spaces, each of which coresponding to one dimension in original action. The sub-actions are presumed to be ordered and interdependent, which as known as \textit{Autoregressive Factorization setting} in \cite{pierrotfactored}. Accordingly, instead of depending on a single value network with a large output layer, we break this into $n$ value networks each with smaller output layers. Figure \ref{pipeline} delineates the separate process. The original action becomes a combination of the $n$ sub-actions. With this methodology, we can achieve fine-grained discretization of individual dimensions and search for the optimal policy within $n$ much smaller action spaces, avoiding an exponentially number of parameters; at the same time, this method is scalable to arbitrarily complex dimensions while maintaining the ability to find the global-optimal policy.  We evaluate MAN on n-dimension maze task, block stacking task and Atari tasks, comparing our model's performance to both DQN and DDQN \cite{van2016deep}. We find that our proposed method outperforms both DQN and DDQN in terms of both the learning speed and the final score. The implementation of MAN is available at \url{https://github.com/keqinw/MAN}.

\section{Related work}

A large amount of the reinforcement learning research focus on the continuous control problems, but less work has been done to address large discrete action spaces. Actually, the number of entities and the size of instances can be very large in many real-world problems, especially in combinatorial research standard problems, where the action spaces may contain thousands of actions. 

Recent works in solving high-dimensional problems usually get influenced by value function approximation, which is one of the common approaches in discrete RL. One of the earliest successful pioneers was TD$(\lambda)$, an online value function approximation algorithm that reached expert-level performance in Backgammon in the early 1900s \cite{tesauro1995temporal}. Q-learning \cite{10.1007/BF00992698}\cite{watkins1992q} is directly derived from TD$(0)$, where for each update step, Q-learning adopts the greedy policy $\underset{a}\max Q (S_{t+1}, a)$. Deep Q-Network (DQN) learning \cite{mnih2013playing}\cite{mnih2015human} was the first work that successfully incorporated deep neural networks with reinforcement learning, creating an algorithm which was able to learn successful control policies for a wide range of classic Atari 2600 video games. A simple improvement to the original DQN algorithm is Double DQN (DDQN) \cite{van2016deep}, which introduces a frozen network to mitigate the over-optimism the single network of DQN accumulates. In this work, we use DDQN for this reason. Dueling DQN \cite{wang2016dueling} is a modification that learns the value $V(s)$ and advantage $A(s, a)$ functions with a single model. Other popular algorithms with similar idea are: Rainbow \cite{hessel2018rainbow}, Fast reward propagation \cite{he2016learning}, Deep Recurrent Q-Network (DRQN) \cite{hausknecht2015deep}, averaged-DQN \cite{anschel2017averaged} and DQV \cite{sabatelli2018deep}.

Recent work assumes certain decompositions in an effort to handle high-dimensional action spaces. As far as we know, Factored Action space Representations (FAR) \cite{sharma2017learning} is the first work that presents the idea of splitting the original action space into smaller sub-actions. Sequential-DQN (SDQN), which was regarded as the first autoregressive algorithm within FARL framework\cite{metz2017discrete}, factorized the N-dimension action into N 1-dimension actions, sharing the same idea with our paper. It leverages two networks $Q_{upper}$ and $Q_{low}$ to predicted the value within original space and the value within factorized spaces. However, they utilizes one $Q_{low}$ network to predict all the 1-dimension actions, which limited the action numbers to be the same in each dimension. Branching Dueling Q-Network (BDQ) \cite{tavakoli2018action} inherites the key idea of Dueling DQN, which firstly evaluates the Q value of original action, then factorizes the action and rate the advantages of each 1-dimension sub-actions. However, since the introduction of advantage, there are more networks needed in learning process. TESSERACT \cite{mahajan2021reinforcement} introduces tenosr decompositions in FARL framework. Amortized Q-learning (AQL) \cite{van2020q} was presented to handle the high-dimensional problems by replacing the costly maximization over all actions with a maximization over a small subset of possible actions drawn from a learned distribution.

In this work, we present a novel improvement to DQN and SDQN: reducing the difficulty of exploration by splitting the original sizeable N-dimension action space into $N$ 1-dimension sub-actions sequentially, inspired by the FARL framework. Each sub-action will be evaluated by a single neural network. This method keeps both simpleness and innovation, and works well in especially large action spaces where alternative DQN-based methods tend to failure, and could be scalable to any complex scenarios, at the same time overcoming the limitations in prior works mentioned above.

\section{Preliminaries}
We consider a Markov Decision Process (MDP), defined by the tuple $(\mathscr{S}, \mathscr{A}, \mathscr{P},r,T,\gamma)$, where $\mathscr{S}$ denotes a set of possible states, $\mathscr{A}$ denotes a set of potential actions, $\mathscr{P}$ denotes the transition dynamics $\mathscr{P}: \mathscr{S} \times \mathscr{A} \times \mathscr{S} \to [0,1]$, $r(s,a)$ as the reward function, $T$ as the task horizon, $\gamma$ as a discount factor $\gamma \in [0,1]$. The goal of reinforcement learning is to search the optimal policy $\pi ^*$ which maximizes the expected discounted reward:
\begin{equation*}
    \pi^*= \underset{\pi}{\arg \max} \mathop{\mathbb{E}_{\rho_\pi}}\left[ \sum^T_{t=0} \gamma^t r\left(s_t,a_t \right) \right]
\end{equation*}
where $\rho_\pi$ is the state-action distribution.

\subsection{Deep Q-networks (DQN)}
Our approach takes its inspiration from DQN \cite{mnih2015human}, which uses a deep Q-network to approximate the high dimensional value functions. Here is the objective function at iteration $i$:
\begin{equation*}
    {L}_i (\theta_i)=\mathop{\mathbb{E}_{s,a,r,s{'}}}\left[\left(y^{DQN}_i-Q(s,a;\theta_i)\right)^2\right]
\end{equation*}
with
\begin{equation*}
    y^{DQN}_i = r + \gamma \max_{a{'}} Q(s{'},a{'};\theta^-_i)
\end{equation*}
where $Q(s{'},a{'};\theta)$ is the Q-network parameterized with $\theta$, and $\theta^-$ represents the parameters of a fixed and separate \textit{\textbf{target network}}. 

This method performs poorly in practice, especially with more complex tasks, such as the Atari 2600 games. To overcome this issue with DQN, researchers created target networks to stabilize the learning by temporally freezing the parameters of the target network $Q(s{'},a{'};\theta^-_i)$, and periodically updating it with the parameters of the main network $Q(s{'},a{'};\theta)$. The specific gradient update is
\begin{equation*}
    \nabla_{\theta_i} {L}_i (\theta_i)=\mathop{\mathbb{E}_{s,a,r,s{'}}}\left[\left(y^{DQN}_i-Q(s,a;\theta_i)\right)\nabla_{\theta_i}Q(s,a;\theta_i)\right].
\end{equation*}

Another key ingredient behind the success of DQN is \textit{\textbf{experience replay}}\cite{lin1992self}. As the agent explores the environment, it populates a replay buffer $D_t=(e_1.e_2,..,e_t)$ with its experiences, where transition $e_t = (s_t,a_t,r_t,s'_t)$. To optimize the networks, transitions are randomly sampled from the replay buffer. The sequence of losses thus takes the form
\begin{equation*}
    {L}_i (\theta_i)=\mathop{\mathbb{E}_{s,a,r,s{'}\sim D}}\left[\left(y^{DQN}_i-Q(s,a;\theta_i)\right)^2\right].
\end{equation*}

This random sampling decorrelates the training data, resulting in more robust training. Additionally, it increases data efficiency by re-using the experience samples.

\subsection{Double Deep Q-networks (DDQN)}
The previous section described the main components of DQN as presented in the work of Mnih \textit{et al.}, 2015. In this work, we use the improved Double DQN (DDQN) learning algorithm of van Hasselt \textit{et al.}, 2016. In Q-learning and DQN, the max operator uses the same values to both select and evaluate an action, which will therefore lead to overly optimistic value estimates\cite{hasselt2010double}. To alleviate this problem, DDQN uses the following evaluation:
\begin{equation*}
    y^{DDQN}_i = r + \gamma  Q(s{'},\underset{a'}{\arg \max}Q(s',a';\theta_i);\theta^-_i)
\end{equation*}

\begin{figure*}[htpb]
  \centering
  \includegraphics[width=\textwidth]{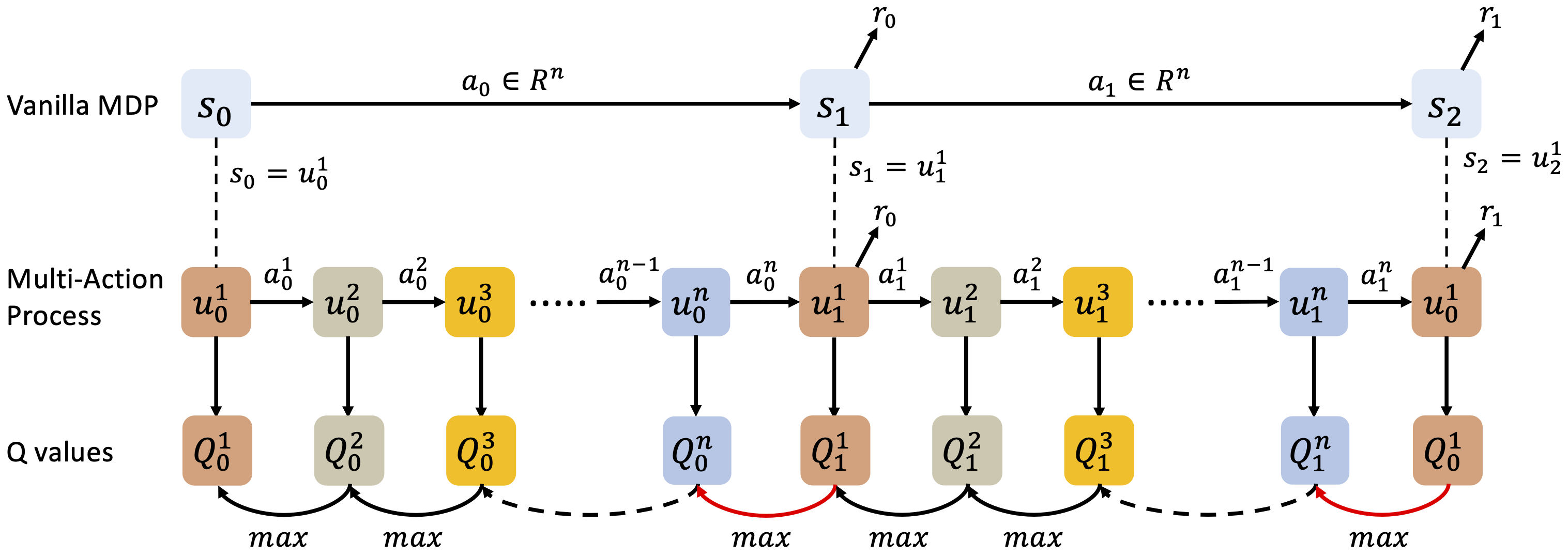}
  \caption{The demonstration of the transition from Vanilla MDP to Multi-Action Process. The first row represents the vanilla MDP, within the n-dimensional space. The second row indicates how the Vanilla MDP may be decomposed into n 1-dimensional sub-actions, by introducing the enrich state $u^k_n$. The superscripts $k$ and subscripts $n$ donate the sub-action index and time-step respectively. The action space within each sub-action could vary with respect to each other, making this decomposition flexible for complex problems. The third row is the Q values that correspond to the sub-actions. In order to connect the Q values together and update them simultaneously, the previous Q value is assign to be equal to the maximization of next Q value, see detail in Definition 3.}
  \label{max_pipeline}
\end{figure*}

\section{MAN: Multi-Action Networks Learning}\label{method}
We now introduce our novel DRL algorithm called Multi-Action Networks (MAN) learning. We first present the online RL algorithm based on TD($\lambda$) together with its update rules, and then extend these update rules to objective functions that can be used to train Artificial Neural Networks (ANNs) for solving complex RL problems.

\subsection{Multi-Action learning based on Q-learning}

The main idea of Multi-Action (MA) learning is to factorize the original action into $n$ sub-actions, thus replacing a larger action space with $n$ smaller 1-d action spaces. For example, in Go, where the original action is 361 coordinate points on the board, we can divide it into two sub-actions, first the horizontal position coordinates(19 actions) and then the vertical position coordinates(19 actions).

The formal definition of MA learning is shown below.

\textbf{Definition 1.}
In this paper, we assume that the action space is factorizable, thus the N-dimension action could be expressed as a Cartesian product of N 1-dimension actions, 
\begin{equation*} 
\mathscr{A} = \mathscr{A}^1 \times \mathscr{A}^2  ... \times \mathscr{A}^n
\end{equation*} 

\textbf{Definition 2.}
\textit{
Multi-Action learning is characterized by:\\
(1) initial action space: $\mathscr{A} = (\mathscr{A}^1,\mathscr{A}^{2}, ..., \mathscr{A}^{n})$\\
(2) n sub-actions: $a^1 \in \mathscr{A}^1,a^{2} \in \mathscr{A}^{2}, ..., a^{n} \in \mathscr{A}^{n}$\\
(3) reward function: $r = r(s,a^1,a^{2}, ..., a^{n})$,\\
(4) enriched state: $u^k_t = (s_t, a^1, a^2, ..., a^{k-1})$,\\
(5) n action values of sub-actions: $Q^{k}(u^k_t,a^{k}_t),k \in n$,\\
(6) transition: $(s_t, a^1_t, a^{2}_t,...,a^{n}_t, r_t, s{'}_t)$.
}

Based on the idea of FARL framework, the original MDP with N-dimension actions is decomposed into the Multi-Action Process which contains sequences of 1-dimension actions. This transition is interdependent. Therefore, as (4) in Definition 2, instead of the original state $s_t$, here we define the enriched state $u^k_t$, which is a concatenation of the state $s_t$ and the history of previous $k-1$ dimension action choices. (Note, $u^1_t = s_t$)


However, problem arises when updating $Q^{n}$ as the action is not fully executed until the agent has executed all $n$ actions $a^{n}$. Thus we cannot apply the objective function mentioned in Section 3.A directly, which connects the reward and q-value in pairs. Therefore, in order to link $a^{n}$ with its `future' reward, we need to add an artificial condition, see as Definition 3.\\
\textbf{Definition 3.} 
\textit{
The value of the previous action equals the maximum value of the next action predicated on this previous action, such that
\begin{equation*} 
    Q^k(u^k_t,a^k_t) = \max_{a^{k+1} \in \mathscr{A}^{k+1}}Q^{k+1}(u^{k+1}_t,a^{k+1}_t),k\in n-1
\end{equation*}
where $u^{k+1}_t = (u^k_t,a^{k}_t) = (s_t,a^{1:k}_t)$.
}

Figure \ref{max_pipeline} demonstrates the Multi-Action Process transition based on the Definition 3. There is a edge case when $k=n$ (shown in red arrow), which is not mentioned in Definition 3. We will introduce this edge case later. We default all the following derivations under the condition $k \in n-1$.

Based on Definition 3, we can get Lemma 1.

\textbf{Lemma 1.}
\textit{
The Q value of $a^{k}$ is the V value of $a^{k+1}$, such that
}
\begin{equation*}
    \begin{aligned}
        Q^k(u^k_t,a^k_t) &= \max_{a^{k+1}\in \mathscr{A}^{k+1}}Q^{k+1}(u^{k+1}_t,a^{k+1}_t)\\
        &= V^{k+1}(u^{k+1}_t)
    \end{aligned}
\end{equation*}

Then we introduce the V value update rule in TD(0):
\begin{equation}\label{v_update}
    V(s_t) := (1-\alpha)V(s_t) + \alpha[r_t + \gamma V(s_{t+1})]
\end{equation}

Based on the Lemma 1, we can rewrite Equation \ref{v_update} by replacing the $V(s_t)$ with $Q(u_t,a_t)$, such that
\begin{equation}\label{update_I}
    Q^k(u^k_t,a^k_t) := (1-\alpha)Q^k(u^k_t,a^k_t) + \alpha[r_t+\gamma Q^k(u^k_{t+1},a^k_{t+1})]    
\end{equation}

The reason we choose V value update rule here, instead of Q value update rule, is that we want to keep the upgrading as simple as possible. V value updating avoids the maximization operation, diminishing the complexity for the future derivation.

Now we have `tentatively' obtained the Equation \ref{update_I} as the value update function for $Q^k$. However, in practice, we find that the model converges very slowly. An alternative module replaces the $Q^k(u^k_{t+1},a^k_t)$ in Equation \ref{update_I} as:
\begin{multline}\label{update_I_new}
    Q^k(u^k_t,a^k_t) := (1-\alpha)Q^k(u^k_t,a^k_t) +\\
    \alpha[r_t+\gamma \max_{a^{k+1} \in \mathscr{A}^{k+1}}Q^{k+1}(u^{k+1}_{t+1},a^{k+1}_{t+1})]   
\end{multline}
where $u^{k+1}_{t+1} = (u^k_{t+1},a^{k}_{t+1}) = (s_{t+1},a^{1:k}_  {t+1})$.

The Equation \ref{update_I_new} is as the same as the Equation \ref{update_I} mathematically according to the Definition 3, nevertheless the practical result shows that Equation \ref{update_I_new} is better. This is perhaps due to the fact that the training with Equation \ref{update_I} is scattered, there is no connection between each $Q^k$. Otherwise, in Equation \ref{update_I_new}, $Q^k$ is always corrected with the next sub-action value $Q^{k+1}$, which strings up the sub-actions as a whole. According to the above view, we chose Equation \ref{update_I_new} to update $Q^k$.

Now let's look back to the edge-case, where $k = n$. Based on Definition 3, Lemma 2 is presented.

\textbf{Lemma 2.}
\textit{The Chain Rule:}
\begin{equation}
    \begin{aligned}
        \max_{a^1 \in \mathscr{A}^1}Q^1(u^1_t,a^1_t) &= \max_{a^1,a^2\in \mathscr{A}^1, \mathscr{A}^2}Q^2((s_t,a^1_t),a^{2}_t)\\
        &= \max_{a^1,a^2, a^3 \in \mathscr{A}^1, \mathscr{A}^2, \mathscr{A}^3} Q^3((s_t,a^1_t,a^2_t),a^{3}_t)\\
        &= ...\\
        &= \max_{ \mathscr{A}} Q^n((s_t,a^1_t,a^2_t,...,a^{n-1}_t),a^{n}_t)\\   
    \end{aligned}
\end{equation}

For better comprehend, here we convert $u^k_t$ back to $(s_t, a^{1:k}_t)$, clarifying the maximization over $a^k_t$. Thinking about the Q value update rule in TD(0) for $Q(u^n,a^n)$:
\begin{multline}\label{q_update}
    Q(u^n_t,a^n_t) := (1-\alpha)Q(u^n_t,a^n_t) + \\
    \alpha[r_t + \gamma \max_{ \mathscr{A}} Q^n((s_{t+1},a^1_{t+1},a^2_{t+1},...,a^{n-1}_{t+1}),a^{n}_{t+1})]
\end{multline}

Use the chain rule to replace the latter part of Equation \ref{q_update}, we can get the update rule for $Q^n$:
\begin{multline}\label{q_update_final}
    Q(u^n_t,a^n_t) := (1-\alpha)Q(u^n_t,a^n_t) + \\
    \alpha[r_t + \gamma \max_{a^1 \in \mathscr{A}^1} Q^1(u^1_{t+1},a^1_{t+1})]
\end{multline}

According to Equation \ref{q_update_final}, we can compensate the edge case for Definition 3 that:
\begin{equation*} 
    Q^n(u^n_t,a^n_t) = \max_{a^{1} \in \mathscr{A}^{1}}Q^{1}(u^{1}_{t+1},a^{1}_{t+1})
\end{equation*}
which marked in red in Figure \ref{max_pipeline}.

So far, we get the update rule for both edge case ($k = n$, Equation \ref{q_update_final}) and none-edge case ($k \in n-1$, Equation \ref{update_I_new}). For the sake of neatness, we combine the two equations into a single final expression by introducing $j$:
\begin{multline}\label{update_final}
    Q(u^k_t,a^k_t) := (1-\alpha)Q(u^k_t,a^k_t) + \\
    \alpha[r_t + \gamma \max_{a^ \in \mathscr{A}^1} Q^j(u^j_{t+1},a^j_{t+1})]
\end{multline}
where $j = 1 + (k \mod n)$.

 
\subsection{Multi-Action Networks learning}

Here we show how to transform the update rules \ref{update_final} as objective functions to train ANNs.

We define neural network $Q^k$ with parameter $\theta$ to approximate the value functions of $a^k$. It is possible to simply express the update rules for MAN-Learning in terms of Mean Squared Error, much like how DQN addresses the updating for Q-Learning.

The \textbf{objective function of $\theta$} is:
\begin{multline}
    {Loss}_{\theta}=\\
    \mathop{\mathbb{E}}[(r_t+\gamma\max_{a^{j}_{t+1}\in \mathscr{A}^{j}}Q^{j}(u^j_{t+1},a^{j}_{t+1})-Q^{k}(u^k_t,a^k_t))^2]
\end{multline}
where $j = 1 + (k \mod n)$.\\


All elements of our novel DRL algorithm have now been defined, which is summarized by the pseudocode presented in Algorithm \ref{algorithm}.
 
\begin{algorithm}
    \caption{MAN Learning for N-dimension Problem}\label{algorithm}
    \begin{algorithmic}[1]
    \State {\bf{Input:}} Set update frequency $F$, minibatch $B$, learning rate $\alpha$, $episode$, dimension of action space $N$;
    \State {\bf{Input:}} N networks $Q^k$ network with parameters $\theta^k$ and ${\theta^-}^{k}$, $k \in N$; $k$ learning errors $\Delta^k = 0$;
    
    \For {$t = 1$ {\bf{to}} $episode$}
        \
        \State Observe $S_t$
        \For {$ p = 1$ {\bf{to}} $N$}
            \State $u^k_t = (s_t,a^{1:k})$, Choose $a^{k}_t \sim {\pi}^{k}(u^k_t)$ 
        \EndFor
        \State Get $R_t,S_{t+1},D_t$;
        \State Store $(S_t, a^1_t, a^{2}_t,...,a^N_t, R_t,S_{t+1},D_t)$ in buffer $D$
        \For {$q = 1$ {\bf{to}} $B$}
            \State Sample transition $v$ from buffer $D$;
            \For {$ k = 1$ {\bf{to}} $N$}
            \State $j = (k \mod N) +1$
            \State $A^{k}_{next} =\underset{a^k}{\arg \max}(Q^{k}(u^k_{q+1};\theta^k))$ 
            \State $y^{k}_q=r_q +(-D_q+1)\gamma Q^{j}(u^j_{q+1},A^{j}_{next} ;{\theta^-}^k)$
            \State $\delta^{\theta}_q = (y^k_q - Q^k(u^k_{q},a^k_q;\theta))^2$
            \State Accumulate change:   
            \State $\Delta^k \gets \Delta^k+ \delta^{\theta}_q \cdot \nabla_{\theta}Q^k(u^k_{q},a^k_q;\theta)/B$
            \EndFor

        \EndFor
        \State Update weights:
        \State For all the $k$, $\theta^k \gets \theta^k + \alpha \cdot \Delta^k$, reset $\Delta^k = 0$
        \If {$t \equiv 0$ mod $F$}
            \State For all the $k$, ${\theta^-}^k := \theta^k$
        \EndIf 
    \EndFor
    
    \end{algorithmic}
\end{algorithm}

\section{Experiments}\label{experiment}
We now show the practical performance of the MAN learning by evaluating our algorithm on a simple n-dimension maze task, a stacking task, and then demonstrating larger scale results for general Atari game-playing.

\subsection{N-dimension Maze}
We firstly exam our algorithm on an n-dimensional Maze task, indicating the MAN is scalable to high dimensions.

\begin{figure}[thpb]
  \centering
  \includegraphics[width=0.48\textwidth]{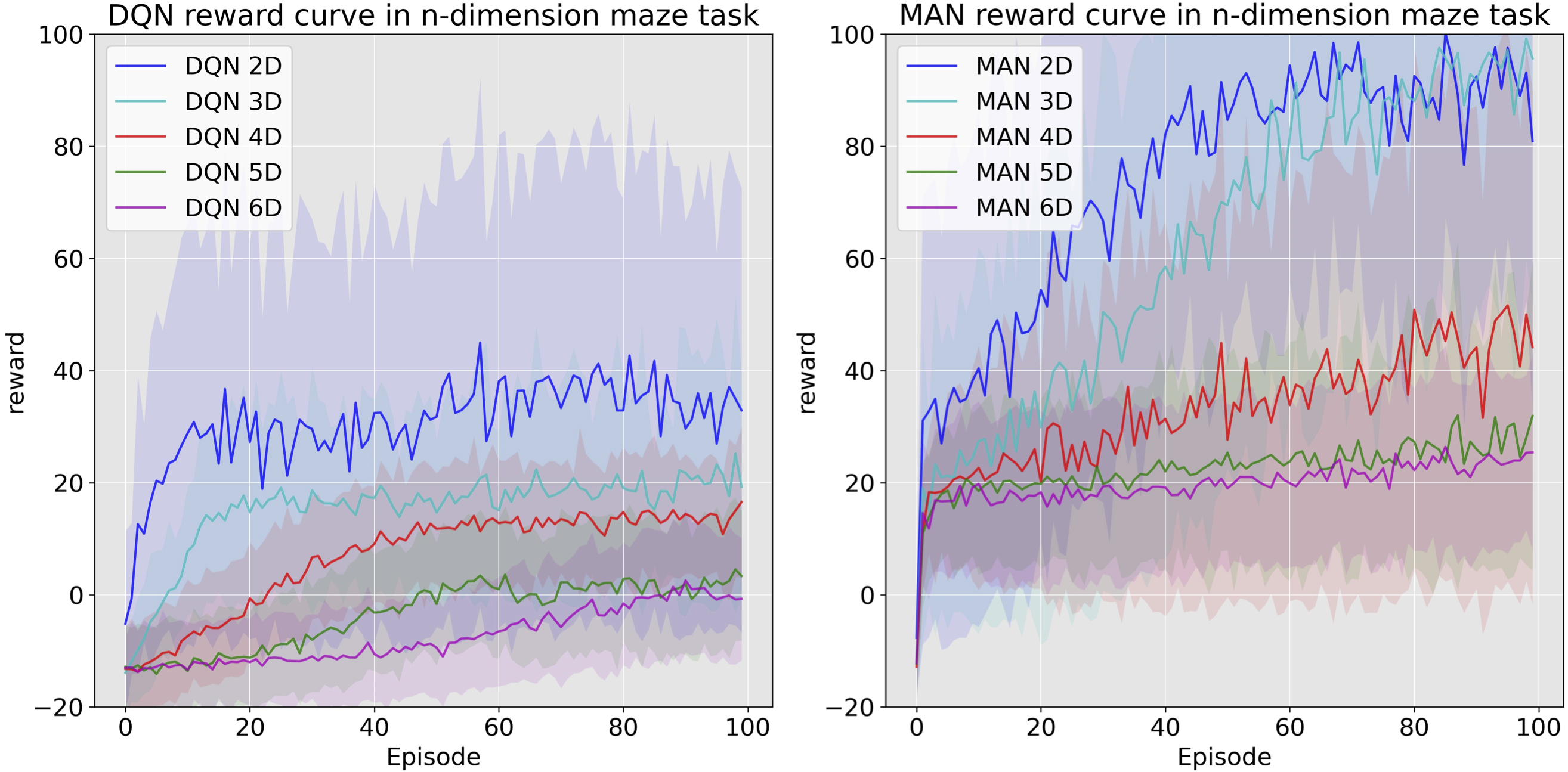}
  \caption{The comparison between MAN and DQN in the N-diemension Maze task. left: DQN, right: MAN. MAN could always learn a better policy, even in the very high dimension setting, such as 6-D.}
  \label{n-dimension result}
\end{figure}

The setting for n-dimension maze is very straight forward: we consider a n-dimension discrete maze task, defined by the tuple $(\mathscr{N},\mathscr{L},\mathscr{S},\mathscr{T}, \mathscr{A})$, where $\mathscr{N}$ denotes the dimension of maze; $\mathscr{L}$ denotes the length of maze, which is equal in each dimension; $\mathscr{S} \in R^n$ is a list which denotes the starting position; $\mathscr{T} \in R^n$ is a list the same as $\mathscr{S}$, denoting the target position; $\mathscr{A}$ denotes the action along each dimension axis. $\mathscr{A}=(-1,0,1)$, corresponding to going forward, no move, and going backward in each dimension respectively. The goal of this task is to touch the target position, departing from the starting position. 

The state $s$ of this task is $(t,c)$ as the combination of target position $t$ and current position $c$. The enriched state $u^k = (s,a^{1:k}),(u^1=s)$.

The size of action space $a \in R^N$ for this task is $3^N$, which grows exponentially and would finally get explosion when the dimension number $N$ is very large. However, in MAN, the original action space is split and sequentialized into $N$ 1-D action spaces $a^k$ with 3 potential action inside. 

The reward function is defined as follow:
\begin{equation*}
  r =
    \begin{cases}
      +100 & \text{if touch the target}\\
      -10 & \text{if stepping out of bounds}\\
      d_{previous} - d_{current} & \text{otherwise}
    \end{cases}       
\end{equation*}
where $d$ is the Manhattan Distance between current position and target position.

In practice, we use $N$ structurally similar networks $Q^k(u^k,a^k), k \in N$ to predict Q value of each 1-D action separately: two hidden layers (256 x 512, 512 x 256) is connected by the activation function (Rectified Linear Unit (RELU)). 

To compare the performance, we implemented DQN, which shares the same internal structure with MAN. Hyper-parameters remain consistent across the two models (DQN and MAN): the discount is 0.99; the greedy-exploration $\epsilon$ linearly decreases from 1 to 0.1 in the first 10000 steps and keeps 0.1 afterwords; the batch size is 32; the soft-update index is 0.005; the learning rates are adjusted to the respective optimal values based on the results of several experiments. 

\begin{figure*}[htpb]
  \centering
  \includegraphics[width=\textwidth]{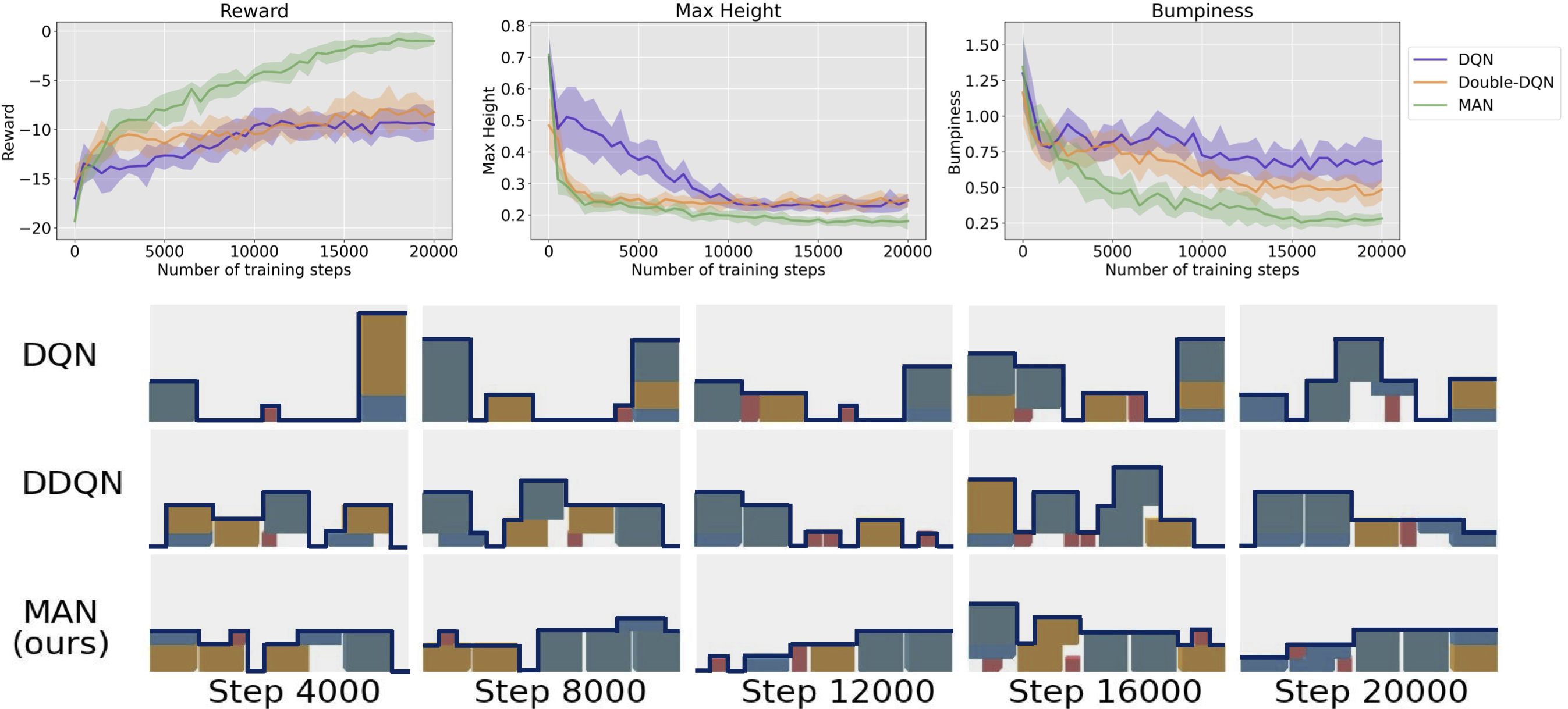}
  \caption{Top: Performances of MAN compared with DQN and DDQN. Bottom: Images of piles captured in testing experiments. The outline of pile is marked in dark blue line.}
  \label{stack_result}
\end{figure*}

In this experiment, we wanted to evaluate our algorithm when changing the dimension number $n$ to observe how performance varied with dimensionality.The results are shown in the Figure \ref{stack_result}. On this experiment, MAN converges both faster and to a larger reward than DQN in all of the dimensionality settings. Although the performance of MAN is still negatively impacted with the increased dimensionality of the maze, our algorithm's performance remains impressive as the action space is very large (729) when the maze dimension is 6.


\begin{figure}[thpb]
  \centering
  \includegraphics[width=0.48\textwidth]{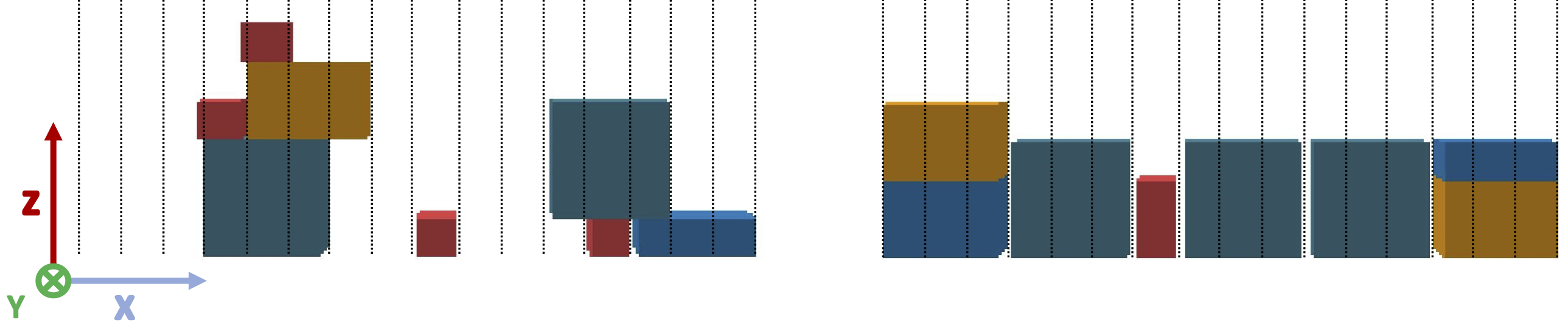}
  \caption{Stacking mission statement: there are four sizes of blocks to choose from. The dotted line divides 16 areas, of which the middle 14 can be used to place blocks. Left: A case of random stacking. Right: A case of tight stacking. Tight means the blocks are compact and the overall height of them are low. }
  \label{stack_intro}
\end{figure}

\subsection{Block Stacking}
We then evaluate the performance of the MAN architecture on a simple real world stacking and packing task to illustrate that our algorithm is suitable to be connected with real world problems.

The goal of this task is to stack rectangular blocks of four different sizes together while making them both as compact as possible and with the lowest overall height possible. Specifically, as shown in Figure \ref{stack_intro}, the block positions vary in the x and z dimensions. We assume blocks have the same width in the y-axis, allowing them to be stacked easily. No rotation is allowed for this simple task. As for observation state, we uniformly discretize the stacking area into 16 positions along x-axis with width 1. The state $S_t$ is defined as the heights of the block outlines in these 16 discrete spaces. As for the stacking action, we need to choose both which type of blocks to stack and where to place the block. Here we design four types of blocks with the sizes of 1x1x1, 1x3x1, 2x3x1 and 3x3x1 respectively. The position selection action sub-space is defined as the middle 14 positions of the stacking area. The leftmost and rightmost positions are not included in this action space since most of blocks in our tasks are of 3 width (if placed in the edge, a part of block will out of environmental bounds). The product of the number of block types (4) and the number of placing positions (14) yields 56 total actions.

Existing algorithms, such as DQN and Double-DQN will have difficulty handling 56 potential actions, because this large action space requires a significant amount of exploration time. However, with MAN this action space is sequentialized and decoupled into block type selection and then block placement, requiring less exploration time and simplifying the task.


In practice, we use two structurally similar networks ($Q^I$, $Q^{II}$) to predict the blocks type ($a^I$) and the placement of blocks ($a^{II}$) separately. The hidden layers and activation function is the same as the N-dimension task.

To compare the performance, we also implemented DQN and DDQN, the network structures of which are similar to that of $Q^I$. The only difference is that it does not partition the action space and therefore the number of neurons in its output layer is 56. The setting of Hyper-parameters remain consistent with the N-dimension Maze task. As for the reward function, we refer to the paper by Junhao Zhang \textit{et al.}. \cite{zhang2020grasp}. The whole process is trained in the Pybullet simulator \cite{coumans2021}.

The results are shown in the Figure \ref{stack_result}. The bumpiness is an indicator calculated by the variance of the outline (marked with blue line in the Figure \ref{stack_result}), the smaller the number, the flatter the top layer of the pile. Within the same training episodes, DQN learned almost no reasonable policy; DDQN learned a good but still imperfect policy; and MAN learned an optimal policy, which outputs the lowest height and bumpiness. 

\subsection{General Atari Game-Playing}

We further evaluated our proposed method on the Arcade Learning Environment (ALE) \cite{bellemare2013arcade}, to confirm MAN's capacity to tackle more complicated problems. The goal is to use a single algorithm and architecture with a defined set of hyper-parameters to learn to play all of the games using only raw pixel observations and game rewards. This environment is widely used as a benchmark in RL due to the high-dimensional observation and the enormous number of diverse games. 


We chose 12 games that combine both button and joystick control, each with an action space of 18 dimensions. Both DQN and DDQN had difficulty mastering joystick and button control at the same time. Also, since the numbers of action in each dimension are different, which are 9 and 2 here, SDQN is impractical in this task. However, MAN could overcome these strains: with MAN, the agent first determines the direction of the joystick (9 actions in total), and then chooses to press or not press the button (2 actions in total). Thus, we reduce 18 actions to 11 (9+2) actions which reduces the exploration need. 

In MAN learning, we use two networks, which are identical except the output layer, to estimate the value of joystick actions and button actions respectively. The network architecture has the same low-level convolutional structure of DQN (Mnih \textit{et al.}, 2015; van Hasselt \textit{et al.}, 2015). The input to the neural network is an $84\times 84 \times 4$ image produced by the pre-process function. The first layer has 32 $8 \times 8$ filters with stride 4, the second layer has 64 $4 \times 4$ filters with stride 2, the third layer has 64 $3 \times 3$ filters with stride 1, and the final layer is a fully-connected layer with 512 units. The output layer is a fully-connected linear layer with the size being the number of valid actions, which is 9 and 2 respectively. We adopt the hyper-parameters of van Hasselt et al. (2015), with the exception of the learning rate which we chose to be slightly higher. Additionally, we chose to use the Adam optimizer \cite{kingma2014adam} to optimize the networks rather than RMSprop \cite{rmsprop}, because it is more empirically stable.

In practice, we closely follow the setup of van Hasselt \textit{et al.} \cite{hessel2018rainbow}, including frame-skipping technique and clipped reward, compare to their results using DQN and DDQN. Additionally, as in \cite{hessel2018rainbow}, we begin the game with up to 30 no-op actions to provide the agent with a random starting position. It is important to note that due to the limitation of computation resources, we only train MAN with 100M steps for each game, which is one-fifth of the original steps in the work of Hasselt \textit{et al.}. The results shown of human performance as well as DQN and DDQN come directly from the work of \cite{van2016deep}.

\begin{figure}[htp]
  \centering
  \includegraphics[width=0.5\textwidth]{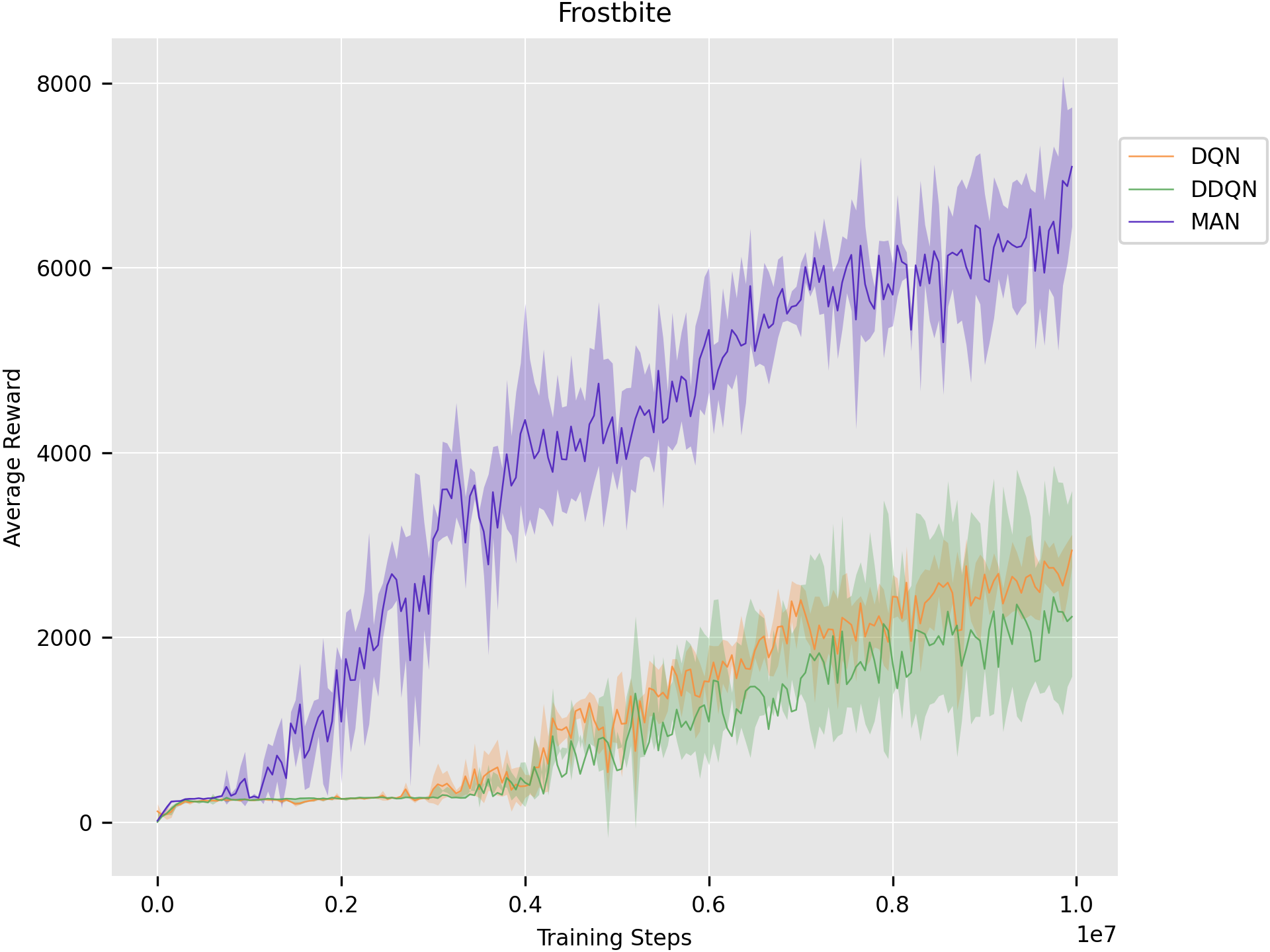}
  \caption{The average rewards by DQN, DDQN and MAN on Frostbite. The results are obtained by running with 6 different random seeds}
  \label{avg_score}
\end{figure}

In order to obtain a uniform standard to assess our method, we calculate the percentage (positive or negative) improvement in score over the best human agent scores using the following equation. 

\begin{equation*}
    \frac{\mathrm{Score}_{Agent} - \mathrm{Score}_{Random}} {\mathrm{Score}_{Human}-\mathrm{Score}_{Random}}  
\end{equation*}

Table \ref{normalized score} lists the normalized scores for each of the 12 games. The best result in each game is marked in bold.


\begin{table}[htp]
\caption{Normalized results for the no-op evaluation condition}
\label{normalized score}
\begin{center}
\begin{tabular}{|l|r|r|r|}
\hline
\textbf{Game} & \textbf{DQN} & \textbf{DDQN} & \textbf{MAN}\\
\hline
\hline
Boxing & 1707.86\% & 1942.86\% & \textbf{2239.29 \%} \\
\hline
Chopper Command & 64.78\% & 42.36\% & \textbf{72.88\%} \\
\hline
Fishing Derby & 95.16\% & \textbf{115.23\%} & 111.11\% \\
\hline
Frostbite & 6.16\% & 4.13\% & \textbf{179.67\%} \\
\hline
H.E.R.O & 76.5\% & 78.15\% & \textbf{79.6}\% \\
\hline
Ice Hockey & \textbf{79.34\%} & 72.73\% & 54.55\% \\
\hline
James Bond & 145.00\% & 108.29\% & \textbf{145.88\%} \\
\hline
Krull & 277.01\% & 350.08 \% & \textbf{989.58\%} \\
\hline
Private Eye & \textbf{2.53\%} & 0.93\% & 0.11\% \\
\hline
Robotank & 508.97\% & 458.76\% & \textbf{581.96\%} \\
\hline
Seaquest & 25.94\% & 39.41\% & \textbf{55.59\%} \\
\hline
Tennis & 143.15\% & 171.14\% & \textbf{265.77\%} \\
\hline
\hline
\textbf{Mean} & 261.0\% & 282.0\% & \textbf{397.7\%} \\
\hline
\textbf{Median} & 87.3\% & 93.2\% & \textbf{128.5\%} \\
\hline
\end{tabular}
\begin{tablenotes}
      \small
      \item \textit{Note}: In above two tables, DQN and DDQN as given by Hasselt \textit{et al.} (2016). The MAN only trained 10 million steps, instead DQN and DDQN trained 50 million steps.
\end{tablenotes}
\end{center}
\end{table}

\begin{figure}[thpb]
  \centering
  \includegraphics[width=0.45\textwidth]{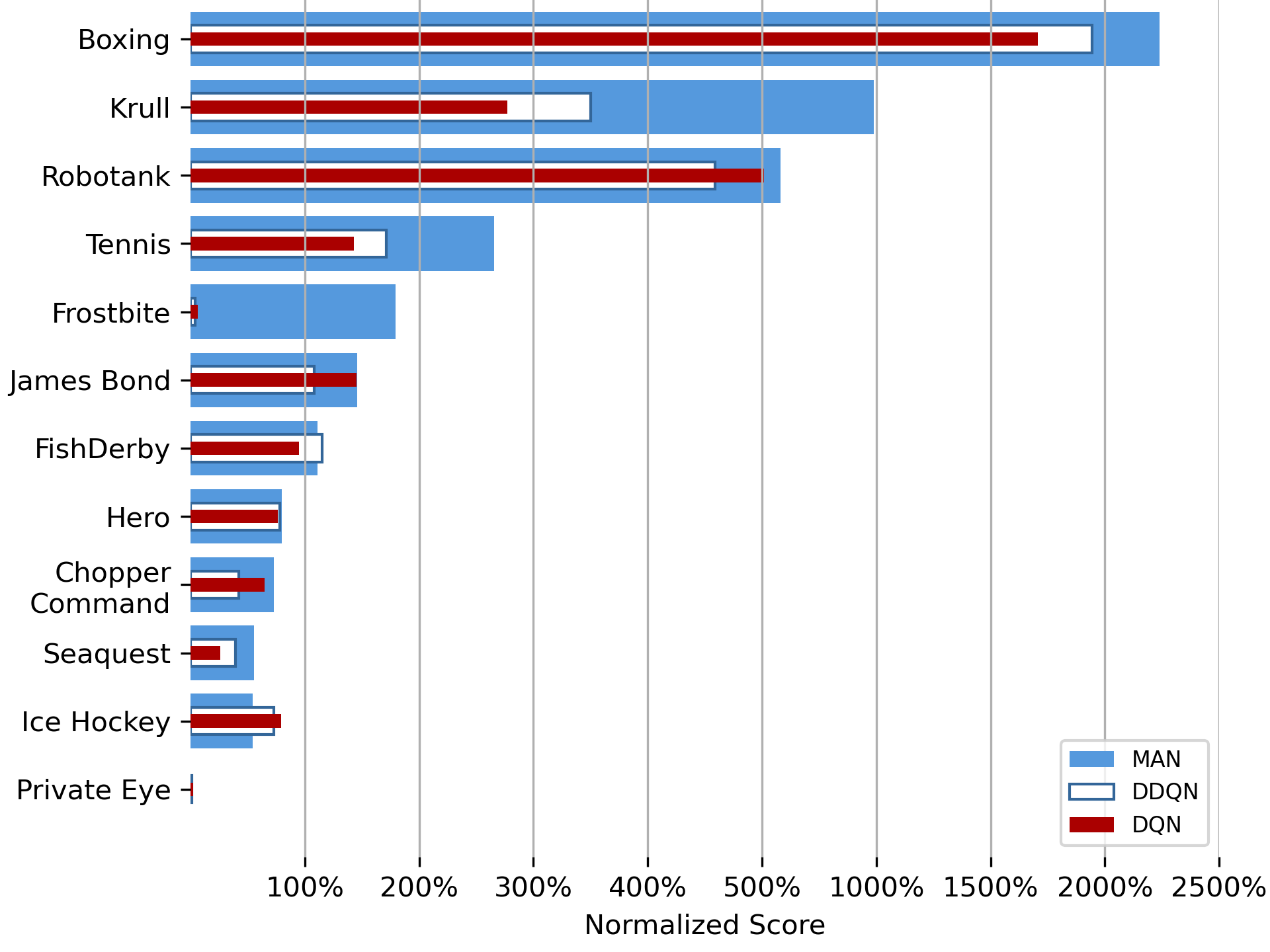}
  \caption{Scores on 12 Atari games were normalized and tested for 20 episodes per game with human starts. The results of DQN and DDQN come from Hasselt \textit{et al.} (2016).}
  \label{atari_result}
\end{figure}

Using this performance measure, it is clear that the MAN learning performs significantly better than both DQN and DDQN. Noteworthy examples include Frostbite (from 6.16\%/4.13\% to 179.67\%), Krull (from 277.01\%/350.08\% to 989.58\%), Seaquest (from 25.94\%/39.41\% to 55.59\%), and Tennis (from 143.15\%/171.14\% to 265.77\%). Figure \ref{avg_score} plots the first 100M steps learning curve in Game Frostbite, which clarifying the significant improvement from MAN. Figure \ref{atari_result} visualizes the improvement of the MAN over the baseline DQN and DDQN. Once again, it needs to be reiterated that this was not a fair game: we only trained for one-fifth of the time compared to the baselines. However, even with this hindrance, we still notice that MAN outperformed the baselines on the vast majority of games. In fact, MAN outperforms DQN in 83.3 percent of games (10 out of 12). It also outperforms the DDQN in 75 percent (9 out of 12) of the games. On 6 of 12 games, our system achieves human-level performance. On most games the improvements from DQN/DDQN to MAN are striking, in many cases bringing scores much closer to human, or even surpassing them. Thus, we assume that if we had been able to train MAN for the same number of timesteps, our results would be even better. 


\section{Conclusion}
We introduced MAN, a novel neural network architecture build off of deep Q-networks that splits the original N-dimension action into N 1-dimension sub-actions, while sharing a common feature learning module. This method, in combination with some algorithmic improvements, leads to dramatic improvements over existing approaches for DRL in the challenging domain of large action spaces. There are two main breakthrough of our approach. Firstly, MAN avoids the exponential increasing with high dimensions, which in turn simplifies exploration with smaller sub-action spaces. Secondly, MAN overcome the limitation in SDQN which could only solve the problem with the same size in each dimension. As what we have exam in Atari 2600 games, our framework is suitable for any combination of discrete actions. 

However, this algorithm does have its limitation: since our key idea lies in the Definition 3, which unites the sub-actions by a maximization relationship, it is unrealistic to deploy MAN in continuous algorithms. Recall some marvelous RL algorithms aim to handle continuous space, such as DDPG\cite{lillicrap2015continuous}, the critic predicts the value of any actions directly, without any maximization operation insides. 

In the future, we plan to expand the usability of our algorithm to more complicated settings, for example, how to use this algorithm in real-world systems. Furthermore, in this work we focused on the simplest decomposition that split the n-dimension action into $n$ 1-d actions. There are other combinatorial structures, such as split the n-d action into a group of 1-d, 2-d, ..., m-d actions ($n = 1+2+...+m$), based on some properties we hope to investigate in the future.
\bibliographystyle{IEEEtran}
\bibliography{ref}

\begin{thebibliography}{10}
\providecommand{\url}[1]{#1}
\csname url@samestyle\endcsname
\providecommand{\newblock}{\relax}
\providecommand{\bibinfo}[2]{#2}
\providecommand{\BIBentrySTDinterwordspacing}{\spaceskip=0pt\relax}
\providecommand{\BIBentryALTinterwordstretchfactor}{4}
\providecommand{\BIBentryALTinterwordspacing}{\spaceskip=\fontdimen2\font plus
\BIBentryALTinterwordstretchfactor\fontdimen3\font minus
  \fontdimen4\font\relax}
\providecommand{\BIBforeignlanguage}[2]{{%
\expandafter\ifx\csname l@#1\endcsname\relax
\typeout{** WARNING: IEEEtran.bst: No hyphenation pattern has been}%
\typeout{** loaded for the language `#1'. Using the pattern for}%
\typeout{** the default language instead.}%
\else
\language=\csname l@#1\endcsname
\fi
#2}}
\providecommand{\BIBdecl}{\relax}
\BIBdecl

\bibitem{mnih2013playing}
V.~Mnih, K.~Kavukcuoglu, D.~Silver, A.~Graves, I.~Antonoglou, D.~Wierstra, and
  M.~Riedmiller, ``Playing atari with deep reinforcement learning,''
  \emph{arXiv preprint arXiv:1312.5602}, 2013.

\bibitem{mnih2015human}
V.~Mnih, K.~Kavukcuoglu, D.~Silver, A.~A. Rusu, J.~Veness, M.~G. Bellemare,
  A.~Graves, M.~Riedmiller, A.~K. Fidjeland, G.~Ostrovski \emph{et~al.},
  ``Human-level control through deep reinforcement learning,'' \emph{nature},
  vol. 518, no. 7540, pp. 529--533, 2015.

\bibitem{dulac2015deep}
G.~Dulac-Arnold, R.~Evans, H.~van Hasselt, P.~Sunehag, T.~Lillicrap, J.~Hunt,
  T.~Mann, T.~Weber, T.~Degris, and B.~Coppin, ``Deep reinforcement learning in
  large discrete action spaces,'' \emph{arXiv preprint arXiv:1512.07679}, 2015.

\bibitem{sutton1999policy}
R.~S. Sutton, D.~McAllester, S.~Singh, and Y.~Mansour, ``Policy gradient
  methods for reinforcement learning with function approximation,''
  \emph{Advances in neural information processing systems}, vol.~12, 1999.

\bibitem{pierrotfactored}
\BIBentryALTinterwordspacing
T.~PIERROT, V.~Mac{\'e}, J.-B. Sevestre, L.~Monier, A.~Laterre, N.~Perrin,
  K.~Beguir, and O.~Sigaud, ``Factored action spaces in deep reinforcement
  learning,'' \emph{submitted to 2021 International Conference on Learning
  Representation (ICLR)}, 2021. [Online]. Available:
  \url{https://openreview.net/forum?id=naSAkn2Xo46}
\BIBentrySTDinterwordspacing

\bibitem{berner2019dota}
C.~Berner, G.~Brockman, B.~Chan, V.~Cheung, P.~D{\k{e}}biak, C.~Dennison,
  D.~Farhi, Q.~Fischer, S.~Hashme, C.~Hesse \emph{et~al.}, ``Dota 2 with large
  scale deep reinforcement learning,'' \emph{arXiv preprint arXiv:1912.06680},
  2019.

\bibitem{sutton1988learning}
R.~S. Sutton, ``Learning to predict by the methods of temporal differences,''
  \emph{Machine learning}, vol.~3, no.~1, pp. 9--44, 1988.

\bibitem{van2016deep}
H.~Van~Hasselt, A.~Guez, and D.~Silver, ``Deep reinforcement learning with
  double q-learning,'' in \emph{Proceedings of the AAAI conference on
  artificial intelligence}, vol.~30, no.~1, 2016.

\bibitem{tesauro1995temporal}
G.~Tesauro \emph{et~al.}, ``Temporal difference learning and td-gammon,''
  \emph{Communications of the ACM}, vol.~38, no.~3, pp. 58--68, 1995.

\bibitem{10.1007/BF00992698}
\BIBentryALTinterwordspacing
C.~J. C.~H. Watkins and P.~Dayan, ``Technical note: \cal q -learning,''
  \emph{Mach. Learn.}, vol.~8, no. 3–4, p. 279–292, may 1992. [Online].
  Available: \url{https://doi.org/10.1007/BF00992698}
\BIBentrySTDinterwordspacing

\bibitem{watkins1992q}
C.~J. Watkins and P.~Dayan, ``Q-learning,'' \emph{Machine learning}, vol.~8,
  no.~3, pp. 279--292, 1992.

\bibitem{wang2016dueling}
Z.~Wang, T.~Schaul, M.~Hessel, H.~Hasselt, M.~Lanctot, and N.~Freitas,
  ``Dueling network architectures for deep reinforcement learning,'' in
  \emph{International conference on machine learning}.\hskip 1em plus 0.5em
  minus 0.4em\relax PMLR, 2016, pp. 1995--2003.

\bibitem{hessel2018rainbow}
M.~Hessel, J.~Modayil, H.~Van~Hasselt, T.~Schaul, G.~Ostrovski, W.~Dabney,
  D.~Horgan, B.~Piot, M.~Azar, and D.~Silver, ``Rainbow: Combining improvements
  in deep reinforcement learning,'' in \emph{Thirty-second AAAI conference on
  artificial intelligence}, 2018.

\bibitem{he2016learning}
F.~S. He, Y.~Liu, A.~G. Schwing, and J.~Peng, ``Learning to play in a day:
  Faster deep reinforcement learning by optimality tightening,'' \emph{arXiv
  preprint arXiv:1611.01606}, 2016.

\bibitem{hausknecht2015deep}
M.~Hausknecht and P.~Stone, ``Deep recurrent q-learning for partially
  observable mdps,'' in \emph{2015 aaai fall symposium series}, 2015.

\bibitem{anschel2017averaged}
O.~Anschel, N.~Baram, and N.~Shimkin, ``Averaged-dqn: Variance reduction and
  stabilization for deep reinforcement learning,'' in \emph{International
  conference on machine learning}.\hskip 1em plus 0.5em minus 0.4em\relax PMLR,
  2017, pp. 176--185.

\bibitem{sabatelli2018deep}
M.~Sabatelli, G.~Louppe, P.~Geurts, and M.~A. Wiering, ``Deep quality-value
  (dqv) learning,'' \emph{arXiv preprint arXiv:1810.00368}, 2018.

\bibitem{sharma2017learning}
S.~Sharma, A.~Suresh, R.~Ramesh, and B.~Ravindran, ``Learning to factor
  policies and action-value functions: Factored action space representations
  for deep reinforcement learning,'' \emph{arXiv preprint arXiv:1705.07269},
  2017.

\bibitem{metz2017discrete}
L.~Metz, J.~Ibarz, N.~Jaitly, and J.~Davidson, ``Discrete sequential prediction
  of continuous actions for deep rl,'' \emph{arXiv preprint arXiv:1705.05035},
  2017.

\bibitem{tavakoli2018action}
A.~Tavakoli, F.~Pardo, and P.~Kormushev, ``Action branching architectures for
  deep reinforcement learning,'' in \emph{Proceedings of the aaai conference on
  artificial intelligence}, vol.~32, no.~1, 2018.

\bibitem{mahajan2021reinforcement}
A.~Mahajan, M.~Samvelyan, L.~Mao, V.~Makoviychuk, A.~Garg, J.~Kossaifi,
  S.~Whiteson, Y.~Zhu, and A.~Anandkumar, ``Reinforcement learning in factored
  action spaces using tensor decompositions,'' \emph{arXiv preprint
  arXiv:2110.14538}, 2021.

\bibitem{van2020q}
T.~Van~de Wiele, D.~Warde-Farley, A.~Mnih, and V.~Mnih, ``Q-learning in
  enormous action spaces via amortized approximate maximization,'' \emph{arXiv
  preprint arXiv:2001.08116}, 2020.

\bibitem{lin1992self}
L.-J. Lin, ``Self-improving reactive agents based on reinforcement learning,
  planning and teaching,'' \emph{Machine learning}, vol.~8, no.~3, pp.
  293--321, 1992.

\bibitem{hasselt2010double}
H.~Hasselt, ``Double q-learning,'' \emph{Advances in neural information
  processing systems}, vol.~23, 2010.

\bibitem{zhang2020grasp}
J.~Zhang, W.~Zhang, R.~Song, L.~Ma, and Y.~Li, ``Grasp for stacking via deep
  reinforcement learning,'' in \emph{2020 IEEE International Conference on
  Robotics and Automation (ICRA)}.\hskip 1em plus 0.5em minus 0.4em\relax IEEE,
  2020, pp. 2543--2549.

\bibitem{coumans2021}
E.~Coumans and Y.~Bai, ``Pybullet, a python module for physics simulation for
  games, robotics and machine learning,'' \url{http://pybullet.org},
  2016--2021.

\bibitem{bellemare2013arcade}
M.~G. Bellemare, Y.~Naddaf, J.~Veness, and M.~Bowling, ``The arcade learning
  environment: An evaluation platform for general agents,'' \emph{Journal of
  Artificial Intelligence Research}, vol.~47, pp. 253--279, 2013.

\bibitem{kingma2014adam}
D.~P. Kingma and J.~Ba, ``Adam: A method for stochastic optimization,''
  \emph{arXiv preprint arXiv:1412.6980}, 2014.

\bibitem{rmsprop}
T.~Tieleman and G.~Hinton, ``Lecture 6.5-rmsprop: Divide the gradient by a
  running average of its recent magnitude.'' \emph{COURSERA: Neural Networks
  for Machine Learning, 4, 26-31}, 2012.

\bibitem{lillicrap2015continuous}
T.~P. Lillicrap, J.~J. Hunt, A.~Pritzel, N.~Heess, T.~Erez, Y.~Tassa,
  D.~Silver, and D.~Wierstra, ``Continuous control with deep reinforcement
  learning,'' \emph{arXiv preprint arXiv:1509.02971}, 2015.

\end{thebibliography}


\end{document}